\documentclass[manuscript]{acmart}
\usepackage{algorithm}
\usepackage{algorithmicx}
\usepackage{amsmath,amsfonts}
\usepackage[noend]{algpseudocode}
\usepackage{textcomp}
\usepackage[caption=false]{subfig}
\AtBeginDocument{%
  \providecommand\BibTeX{{%
    \normalfont B\kern-0.5em{\scshape i\kern-0.25em b}\kern-0.8em\TeX}}}

\setcopyright{acmcopyright}
\copyrightyear{2022}
\acmYear{2022}

\begin{document}

\title[Vision Checklist]{Vision Checklist: Towards Testable Error Analysis of Image Models to Help System Designers Interrogate Model Capabilities}

\author{Xin Du}
\email{x.du@ed.ac.uk}
\affiliation{%
  \institution{University of Edinburgh}
  \country{United Kingdom}}

\author{Benedicte Legastelois}
\email{blegaste@ed.ac.uk}
\affiliation{%
  \institution{King College of London}
  \country{United Kingdom}}

\author{Bhargavi Ganesh}
\email{b.ganesh@sms.ed.ac.uk}
\affiliation{%
  \institution{University of Edinburgh}
  \country{United Kingdom}}

\author{Ajitha Rajan}
\email{arajan@staffmail.ed.ac.uk}
\affiliation{%
  \institution{University of Edinburgh}
  \country{United Kingdom}}

\author{Hana Chockler}
\email{hana.chockler@gmail.com}
\affiliation{%
  \institution{King College of London}
  \country{United Kingdom}}


\author{Vaishak Belle}
\email{vbelle@ed.ac.uk}
\affiliation{%
  \institution{University of Edinburgh}
  \country{United Kingdom}}

\author{Stuart Anderson}
\email{soa@staffmail.ed.ac.uk}
\affiliation{%
  \institution{University of Edinburgh}
  \country{United Kingdom}}

\author{Subramanian Ramamoorthy}
\email{s.ramamoorthy@ed.ac.uk}
\affiliation{%
  \institution{University of Edinburgh}
  \country{United Kingdom}
}








\renewcommand{\shortauthors}{Xin Du etc.}

\begin{abstract}
Using large pre-trained models for image recognition tasks is becoming increasingly common owing to the well acknowledged success of recent models like vision transformers and other CNN-based models like VGG and Resnet. The high accuracy of these models on benchmark tasks has translated into their practical use across many domains including safety-critical applications like autonomous driving and medical diagnostics. Despite their widespread use, image models have been shown to be fragile to changes in the operating environment, bringing their robustness into question. There is an urgent need for methods that systematically characterise and quantify the capabilities of these models to help designers understand and provide guarantees about their safety and robustness. In this paper, we propose Vision Checklist, a framework aimed at interrogating the capabilities of a model in order to produce a report that can be used by a system designer for robustness evaluations. This framework proposes a set of perturbation operations that can be applied on the underlying data to generate test samples of different types. The perturbations reflect potential changes in operating environments, and interrogate various properties ranging from the strictly quantitative, e.g., robustness to dropped patches, to more qualitative, e.g., robustness to texture and colour variations. Our framework is evaluated on multiple datasets like Tinyimagenet, CIFAR10, CIFAR100 and Camelyon17 and for models like ViT and Resnet. Our Vision Checklist proposes a specific set of evaluations that can be integrated into the previously proposed concept of a model card. Robustness evaluations like our checklist will be crucial in future safety evaluations of visual perception modules, and be useful for a wide range of stakeholders including designers, deployers, and regulators involved in the certification of these systems. Source code of Vision Checklist would be open for public use.
\end{abstract}



\maketitle

\section{Introduction}
Recent concerns in trustworthy AI suggest that machine learning models need to be rigorously tested before being deployed in safety-critical applications~\cite{wing2021trustworthy}. To increase  transparency in the model generation, testing, and evaluation processes. Mitchell, et.al~\cite{mitchell2019model} proposed the creation of model cards, which are documents that contain performance characteristics for a given model, and details on how the model was intended to be used\cite{mitchell2019model}. As public concern about the unintended consequences of machine learning models grows, there is an increasing pressure to introduce certification criteria for systems that use these models. This has already been translated into preliminary regulatory guidance in domain areas like medicine~\cite{belle2021principles}. As part of the UK Medicines and Healthcare products Regulatory Agency (MHRA) 'Software and AI as a Medical Device Change Programme', the MHRA emphasises the need for "robust assurance with respect to safety and effectiveness" of medical devices, adding the need for assurance that they "function as intended"~\cite{noauthor_software_nodate}. Similarly, the US Food and Drug Administration (FDA)'s proposed approach notes that "the FDA would expect a commitment from manufacturers on transparency and real-world performance monitoring for artificial intelligence and machine learning-based software as a medical device"~\cite{health_artificial_2021}.

The preliminary guidance above provides concrete examples of cases where there would be benefits to creating standardised evaluation criteria for machine learning models. In their model cards framework, Mitchell, et.al specifically point out the need for thorough model evaluations, noting that “evaluation datasets should not only be representative of the model’s typical use cases but also anticipated test scenarios and challenging cases”~\cite{mitchell2019model}. This problem is referred to in the theoretical literature as the generalization of machine learning models, and is considered the primary goal of learning theories~\cite{vapnik1994measuring}. The general methodology for testing the generalization ability of a model is to use a small set of hold-out data. Even though this methodology is useful, this hold-out data  may not be fully representative of the deployment environment, due to false parameter estimation, distribution shift, unknown distribution etc. The hold-out data may therefore introduce biases such as over-confident performance~\cite{ribeiro2020beyond}. This problem indicates that a single aggregated statistical measure is not enough to evaluate the complex performance outcomes of machine learning models. 

This paper presents the Vision Checklist tool which allows for comprehensive evaluation of image classification models. The checklist assesses model performance in response to specific image perturbations and test types. Our analysis focuses on vision transformer based image models like ViT and DeiT~\cite{dosovitskiy2020image,touvron2021training}. We focus on these models due to their recent success and increasingly widespread use. The transformer structure has proven to be highly successful in NLP tasks~\cite{vaswani2017attention}, and recently, the vision transformer structure has outperformed the SOTA across the main benchmarks. There is a lot of work being done to assess and explain image models, such as evaluating against adversarial attacks~\cite{goodfellow2014explaining,madry2018towards}, determining fairness of image recognition~\cite{wang2020towards,buolamwini2018gender}, localized saliency region explanation~\cite{selvaraju2017grad}, counterfactual generation~\cite{chang2018explaining}, and determining robustness on out-of-distribution (OOD) data~\cite{qin2021understanding,du2021beyond,zhang2021out}. However, existing work focuses on a single specific task like counterfactual generation, or robustness on OOD data rather than addressing several evaluation tasks at once. To meet the needs of stakeholders, including designers, deployers, regulators, and users, while matching the potential deployment environment, model evaluations need to be more comprehensive. To that end, Vision Checklist generates a set of test samples for different model evaluation tasks like fairness, robustness, counterfactual analysis, and formulates a report on a vision model's performance in response to all these tasks. In this paper, we start by applying our Vision Checklist on vision transformer models, and then compare the results to those generated by using CNN structure based image models such as VGG and Resnet~\cite{he2016deep,simonyan2014very}. We envision the Vision Checklist as being used within a model card, as a standard evaluation criteria for vision models~\cite{mitchell2019model}.

\begin{figure}[t]
\includegraphics[width=.8\columnwidth]{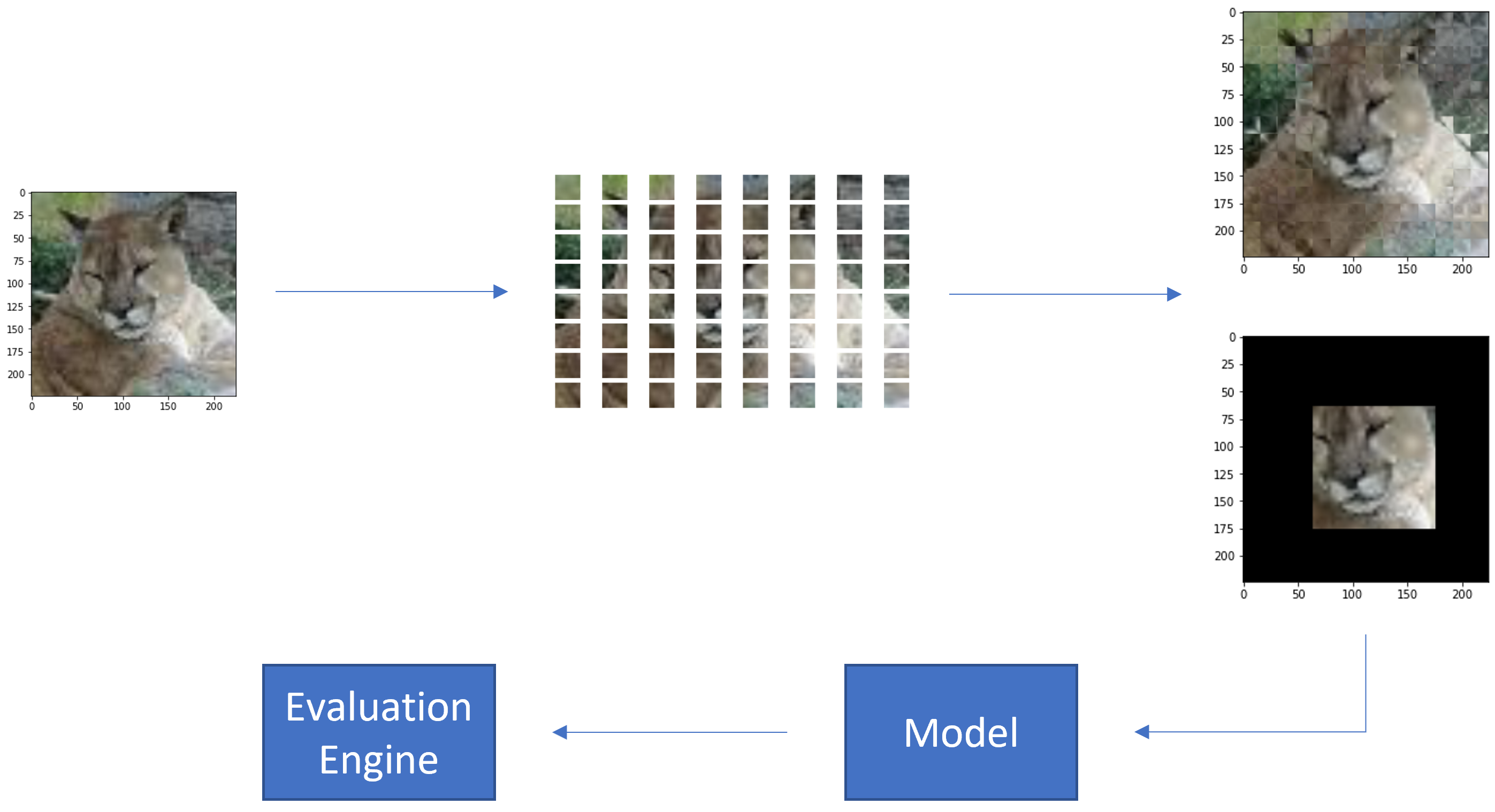}
\caption{Framework for Vision Checklist}
\label{fig:framework}
\end{figure}

\begin{figure}[t]
 \centering
   \subfloat[An image of train\label{fig1}]
   {\includegraphics[width=0.19\columnwidth]{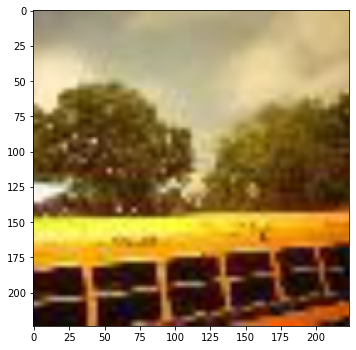}}
   \hspace{1mm}
   \subfloat[Patch rotation\label{fig2}]
   {\includegraphics[width=0.19\columnwidth]{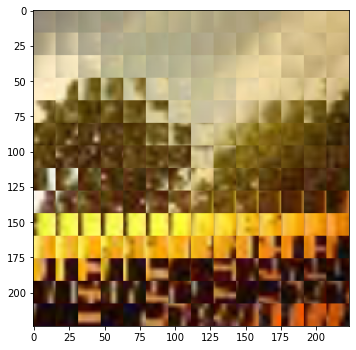}}
   \hspace{1mm}
   \subfloat[A bird on land background\label{fig3}]
   {\includegraphics[width=0.24\columnwidth]{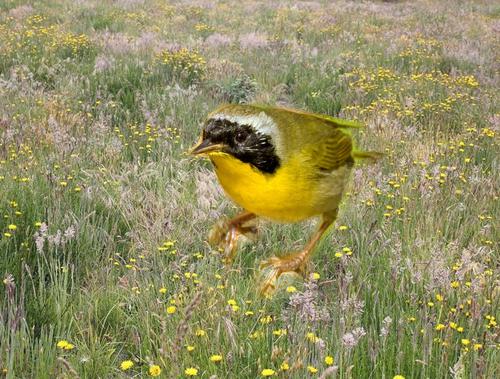}}
   \hspace{1mm}
   \subfloat[A bird on water background\label{fig4}]
   {\includegraphics[width=0.24\columnwidth]{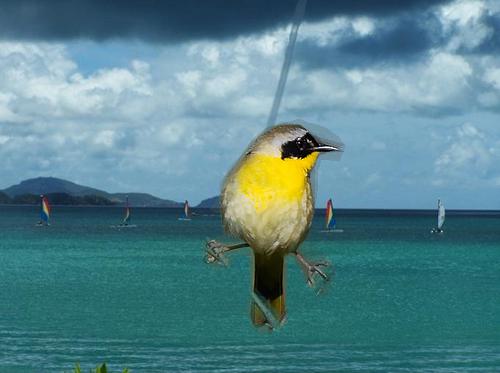}}
   \caption{Perturbed image examples for different tests.}
    \label{fig:2}
\end{figure}

In each of the test cases generated by Vision Checklist, we break down the potential failures of the model into different categories. Instead of directly manipulating image pixels, the basic elements of our image manipulation operator are image patches (cf. \@ Figure \ref{fig:framework}). In Figure~\ref{fig1},\ref{fig2}, we generate the test sample by randomly rotating image patches. This is to test whether the model learns non-robust features when the semantic information of the image has been corrupted. In Figure~\ref{fig3},\ref{fig4}, we generate a test sample by replacing the background of the images. This is to test whether the model can correctly predict the target label under distributional shifts. 

As an example, Figure~\ref{fig:f1g1} shows an original image with the label `cougar', and a series of samples $\tilde{x}$.
Here we choose the vision transformer model as the model to evaluate~\cite{dosovitskiy2020image}. 
The image model $f(\cdot)$ predicts the correct label for the original image. Then we generate different test samples with different operators. In Figure~\ref{fig:f1g1}, we generate test samples by random 50\% rotation of patches, and rotating patches only in the central region of the image. The generated $\tilde{x}$ is passed into model $f$ to test whether the result is the same as the expected output. In Figure~\ref{fig:f1g3}, we generate a test sample by shuffling the image patches. The operation breaks the semantic information in the original image, which changes the label determined by the model. We also generate test samples using counterfactual perturbation by first using the model under test to identify the important regions, and then infilling those regions to create a perturbed image. This perturbed image is passed into the model to investigate whether it can predict the correct label. 


\section{Considerations for model evaluation systems}
In this section we identify properties we believe should be exhibited by any good model evaluation system. Error analysis plays an important role in model evaluation. Analyzing error modes under different situations provides insight into the capabilities of the model at hand, and ensures that the model performance is better understood by designers and developers. However, current evaluation systems' analysis of the sources and types of errors is unsystematic and insufficient. We propose four properties of model evaluation systems that tackle defects in error analysis: faithfulness, sufficiency, necessity, and identifiability.

\subsection{Faithfulness}
Faithfulness requires that model evaluation should be reproducible~\cite{wu2019errudite}. Error rates and error modes generated by a model evaluation system should be replicable relative to a set of hypotheses, i.e. produce the same results when another set of experiments are conducted with the same hypotheses. Before generating test samples, it is important to precisely define the hypothesis being used. For instance, patch segmentation is an essential component for a vision transformer. In order to investigate the reaction of a vision transformer to patch based rotation, we could define the following hypothesis: `under 50\% patch rotation by random selection, the semantic ground truth is retained from the original images.' Here there are three meanings that can be gleaned from this hypothesis. First, the test samples are generated by applying a rotation operation on existing images. Second, the image patches to be rotated are randomly selected. Third, the semantic ground truth is inherited from the original images. An accurately defined hypothesis should always be related to the semantic meaning of the data. For example, the MNIST dataset is generated under the hypothesis that humans first have an image of the digit in their minds, and then write down the digit to formulate an image~\cite{scholkopf2012causal}.  If one were using the MNIST dataset to generate test samples, faithfulness would require any hypothesis to match the anti-causal relationship between images and semantic meanings present in the underlying data.

\subsection{Sufficiency}
The general method for validating a deep learning model is to test its performance on a small amount of held-out data. However, this is not enough to recover all error types. Sufficiency requires that a  model evaluation system should consider the entire dataset as candidates for test samples. Here by 'entire' we mean training, validation, and testing data. In addition, a sufficient evaluation system should use not only  in-distribution data, but also use out-of-distribution data. 
As mentioned by~\cite{wu2019errudite}, evaluation systems that use a small subset of test data can only recover 5~\% to 10~\% errors compared with those that use the entire dataset. Considering the entire dataset as candidates for generating test samples therefore allows us to create a more scalable evaluation system.

\subsection{Necessity}
Error analysis plays an important role in model evaluation systems. However, a good evaluation system should not overfit errors. The necessity principle requires that both error rates and correct predictions should be considered. Due to Simpson's Paradox~\cite{hernan2011simpson}, performance metrics like accuracy in might demonstrate contradictory results across subgroups~\cite{duivesteijn2016exceptional,du2020fairness}. Hence, data grouping techniques are essential for model evaluation. As shown by~\cite{sagawa2020investigation}, overparameterized deep neural networks can achieve high accuracy on average for the test dataset, but generate a low worst case group accuracy. In this case, necessity requires that the evaluation system should focus both on the error mode in groups where the model produced wrong predictions, and on the groups where the model produces mostly correct predictions. The idea would be to analyze the similarity between groups and use this comparison to come up with an overall evaluation of the model's performance.

\subsection{Identifiability}
Identifiability is an important concept in statistics and probability theory. Here we refer to structural identifiability~\cite{bellman1970structural}. For a given model and observations, if it is possible to learn a unique parametrization for that model, then we say it is identifiable. Identifiability is essential for ensuring the replicability and interpretability of a model~\cite{brunner2019identifiability}; both of which contribute to the quality of a model evaluation system. In this paper, we define \textit{identifiability} as the following: for a given hypothesis setting, if there are two groups of outputs corresponding to that setting, then we say it is unidentifiable.

\section{Vision Checklist}
Before applying our Vision Checklist to specific examples, we need to define the set of transformation operators that we use. The basic functions used to perturb images are
rotation, blur, and shuffle. After defining the transformation operators, we define a collection of test types. Each test type aims to evaluate different model capabilities and each has a set of associated hypotheses that are defined using the properties of model evaluation systems elaborated on in Section 2.

\subsection{Transformation Operators}

\subsubsection{Rotation}
\begin{figure}[t]
\centering
\includegraphics[width=0.6\columnwidth]{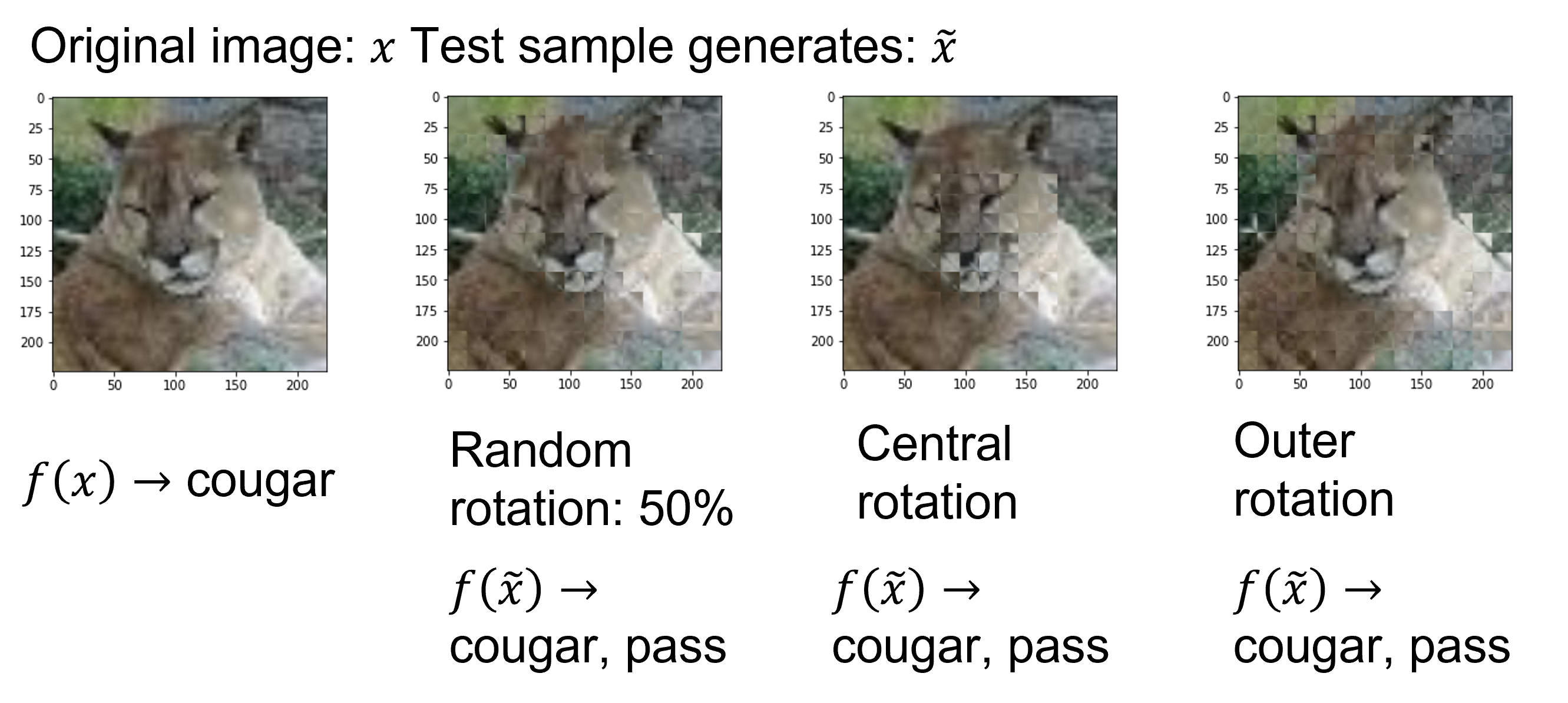}
\caption{Test examples on Rotation operator.}
\label{fig:f1g1}
\end{figure}

This operator changes the angles of image patches, to break the semantic information to some degree. For example, in Figure~\ref{fig:f1g1}, three test examples are generated using the rotation operator. The semantic label of the generated images is still recognizable to a human, and the model predicts the correct label.

\subsubsection{Occlude}
\begin{figure}[t]
\centering
\includegraphics[width=0.6\columnwidth]{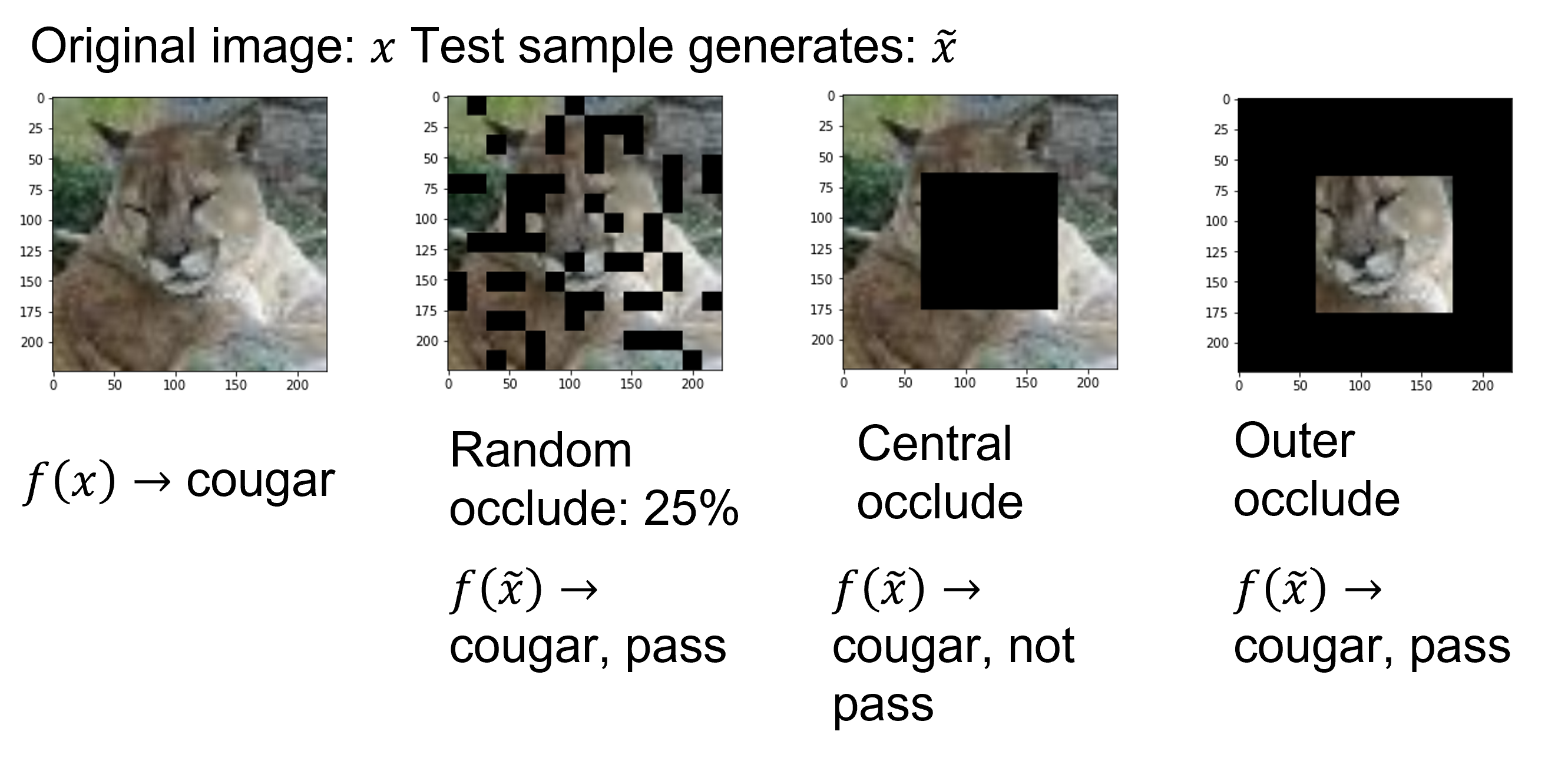}
\caption{Test examples on occlude operator.}
\label{fig:f1g2}
\end{figure}

The occlude operator replaces image patches with uninformative background in order to perturb the semantic information of the images. For example, in Figure~\ref{fig:f1g2}, three test examples are generated using the occlude operator. Where the center is occluded, an important part of this image has been replaced with uninformative background. However, the model still predicts the correct label. For other generated test samples, the model predicts the correct label despite the image perturbations.

\subsubsection{Blur}
The blur operator downsamples the image to limited pixel density, and substitutes the downsampled image to reconstruct the image.

\subsubsection{Shuffle}
For the shuffle transformation, the input image is separated into image patches and then reconstructed as a patch sentence. The shuffle operator permutes the image patches in the original patch sentence. Figure~\ref{fig:f1g3} illustrates this.  In this example, the operator changes the label that the model assigns to the image.

\subsection{Test Types}
The basic workflow for model testing is that for a given test sample, according to a presumed hypothesis, a certain operation is applied to perturb the image and generate the test sample. Then the test sample is passed into the model, which generates the prediction. The prediction is evaluated against the hypothesis to generate the test report. In this section, following this workflow, we define different test types using various hypotheses, to test the different capabilities of the model.


\subsubsection{Random perturbation examples}
In this test type, the patches are selected by randomly sampling from the image. A Bernoulli distribution is presumed to simulate a binary selection process for each patch. This test treats the contribution of each patch as the same, and a unified statistic is applied to give a final evaluation of the model's performance. This test aims to evaluate how the model would perform against random perturbation of the images.

\subsubsection{Central and outer perturbation examples}
In this test type, the hypothesis is that the central part of the image usually makes more important contributions to the label. The transformation operations are applied to either the central or the outer part of the patches. The aim of this test is to evaluate how the model would perform against perturbations of important and unimportant regions.

\subsubsection{Adversarial examples}
\begin{figure}[t]
\centering
\includegraphics[width=0.6\columnwidth]{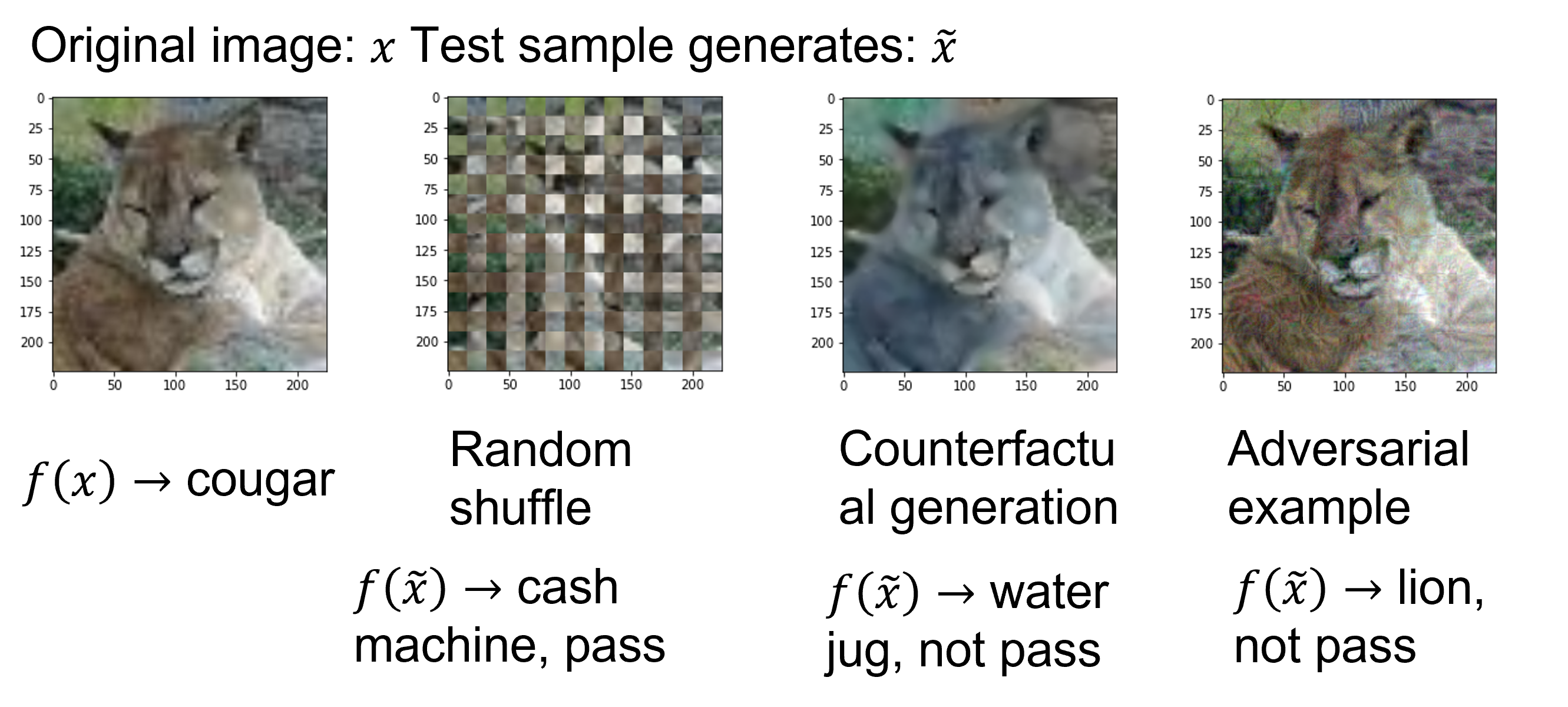}
\caption{Test examples on shuffle, counterfactual and adversary.}
\label{fig:f1g3}
\end{figure}

The aim of adversarial attacks~\cite{madry2018towards} is to find samples that are close to a sample $x$ but are labelled differently by the model than the label given to $x$. In this test type, we employ a PGD attack~\cite{madry2018towards} to generate adversarial examples for each input image. As shown in Figure~\ref{fig:f1g3}, the model output for the adversarial example is 'lion' while the ground truth is 'cougar'. This case also shows that the model has suffered from texture bias~\cite{hermann2019origins}. The generating of adversarial examples is model-based and does not involve the basic transformation operators. Hence, each time we run this test, the algorithm utilizes projected gradient descent to find optimized adversarial examples.

\subsubsection{Counterfactual examples}
In this test type, rather than select patches randomly or based on central assumptions, we use an algorithm~\cite{chang2018explaining} to detect the regions that are important for model prediction. The important regions are infilled by methods like Gaussian mean sampling. Depending on the results of infilling, the generated images either remain recognizable to the human eye, and maintain the same label, or contain a new label. Then the generated examples are passed into the model to compare the prediction with the ground truth.

\subsubsection{Out-of-distribution examples}
Recently, out-of-distribution(OOD) generalization~\cite{zhang2021out} has been a popular topic. There are different definitions of OOD generalization. In this paper, we investigate this problem from the perspective of distributional shifts. The aim of this test is to evaluate whether the model under evaluation focuses on core features or spurious features. In this context, by spurious features, we mean features like background, brightness, color etc. In Figure~\ref{fig3},\ref{fig4}, the backgrounds of the images are replaced. If the model learns spurious correlation with the background information, then the prediction might change.

\subsubsection{Visual Explanations}
\begin{figure}[t]
 \centering
   \subfloat[Positive\label{fig:vis1}]
   {\includegraphics[width=0.2\columnwidth]{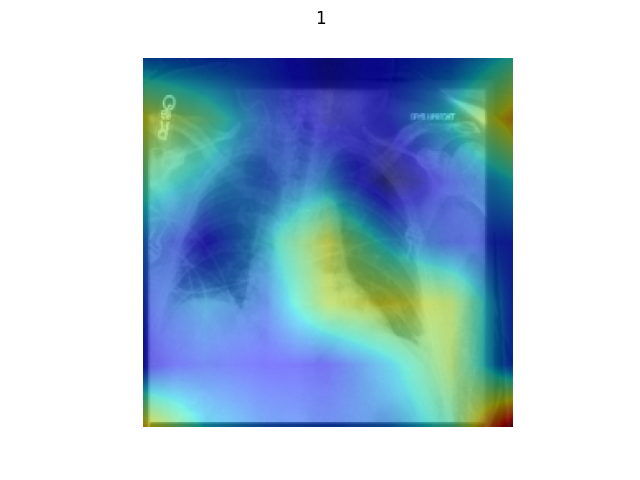}}
   \subfloat[Positive\label{fig:vis2}]
   {\includegraphics[width=0.2\columnwidth]{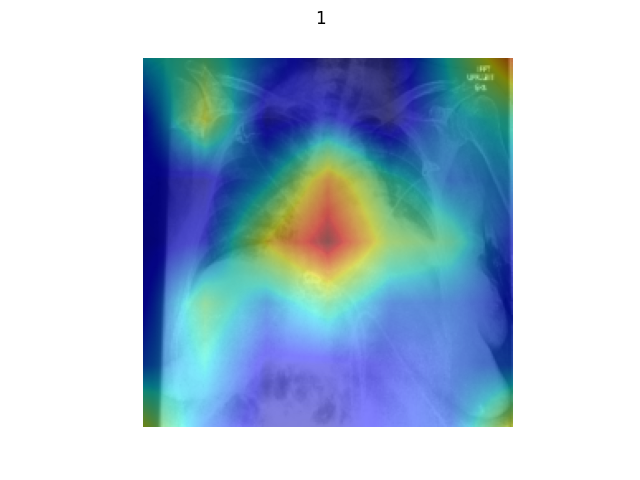}}
   \subfloat[Negative\label{fig:vis3}]
   {\includegraphics[width=0.2\columnwidth]{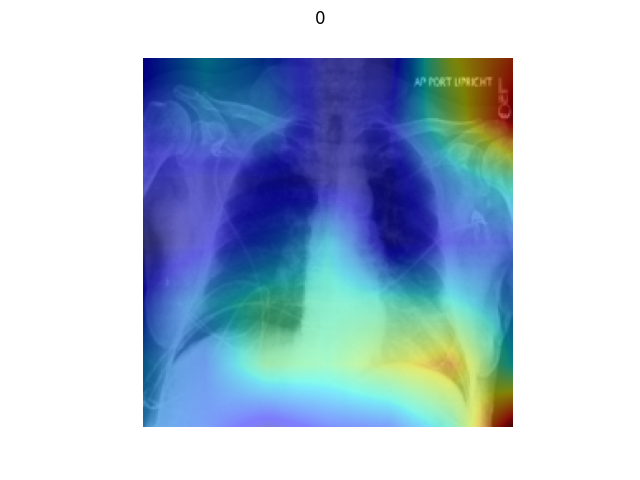}}
   \subfloat[Negative\label{fig:vis4}]
   {\includegraphics[width=0.2\columnwidth]{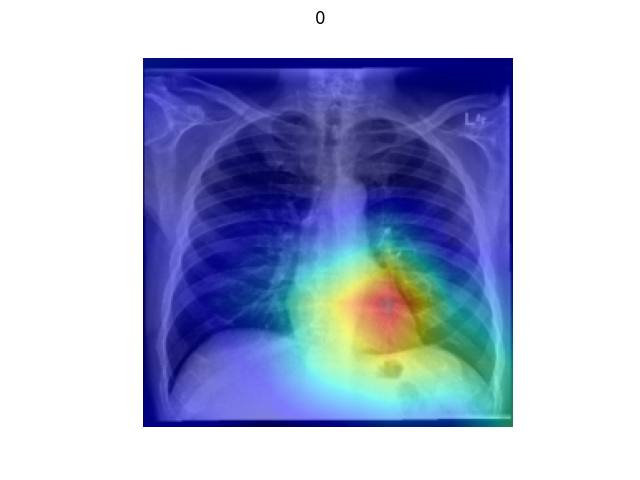}}
   \caption{Visual explanation for Resnet50 on Chexpert dataset. Label indicates the appearance of `Atelectasis'.}
    \label{fig:chexpert}
\end{figure}

As a special test type, visual explanation tools can  provide richer information for evaluating the performance of the model than pure error rates. As discussed by~\cite{selvaraju2017grad}, visual explanations can help users identify error modes, gain trust when establishing and deploying models, and learn more about the model based on its predictions. In Figure~\ref{fig:chexpert}, we demonstrate several visualizations for Resnet50 on the Chexpert dataset using the Grad-CAM method. The test here is conducted based on the hypothesis that 'perturbation by the basic operators in low contribution regions change the label with low probability AND perturbation by the basic operators in the high contribution regions change the label with high probability.' With this hypothesis, users can evaluate the correctness of the prediction using their subjective cognition of semantic meaning.

The list of transformation operators and test types does not have to be limited to the ones presented here. For example, for some models it may be appropriate to consider tests that include recolouring, scaling, translation, etc. The operators and test types used can be varied depending on the domain of interest.

\subsection{Automatic generating process}
In order to apply the previous transformations and test types, a user can either generate test examples from scratch or consider more automated approaches. For some situations, datasets created from scratch with strong prior information are useful for testing. For instance, the out-of-distribution generalization test type requires the creation of datasets by varying spurious features. In Figure~\ref{fig3} and \ref{fig4}, based on the hypothesis that background is a spurious feature, the background is manually replaced to create test samples. Creating such test data from scratch could accurately address the aim of checklist testing. However, this approach is time consuming and would not scale to large datasets. 

\begin{figure}[t]
 \centering
   \subfloat[rotation base image \label{frr}]
   {\includegraphics[width=0.2\columnwidth]{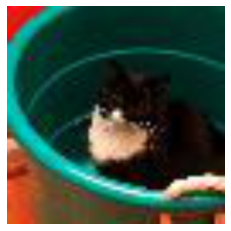}}
   \subfloat[random rotation 25\% \label{frr025}]
   {\includegraphics[width=0.2\columnwidth]{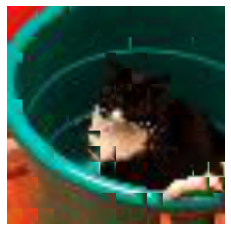}}
   \hspace{1mm}
   \subfloat[random rotation 50\% \label{frr050}]
   {\includegraphics[width=0.2\columnwidth]{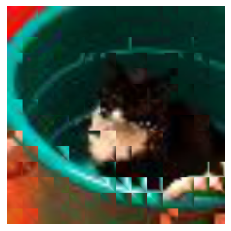}}
   \hspace{1mm}
   \subfloat[random rotation 75\% \label{frr075}]
   {\includegraphics[width=0.2\columnwidth]{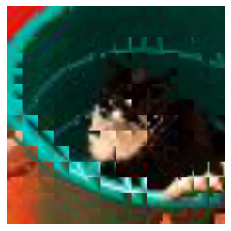}}
   \caption{Templates for rotation operation}
    \label{fig:tpro}
\end{figure}

\begin{figure}[t]
\centering
\includegraphics[width=0.6\columnwidth]{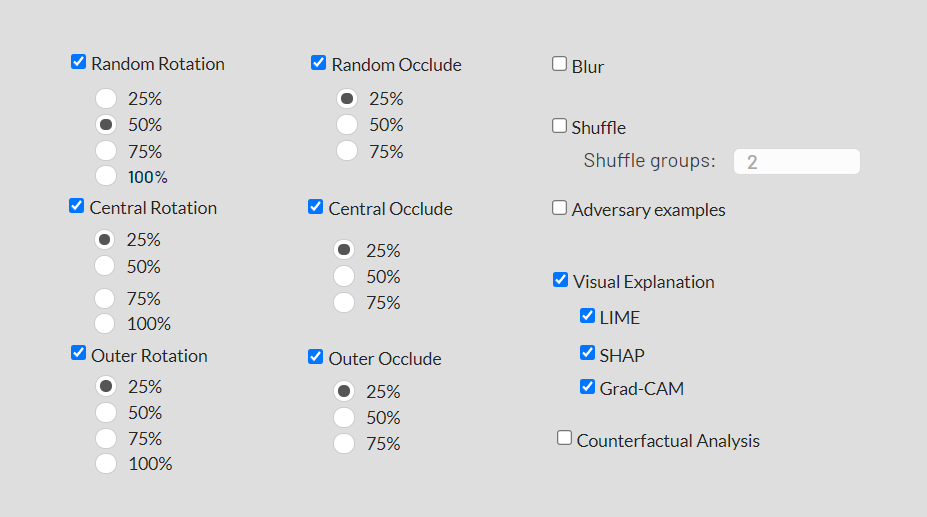}
\caption{User interface for the test sample generating template.}
\label{fig:template}
\end{figure}

In order for the approach to scale, we need to begin by selecting a reasonably large collection of test images. Using this collection as a starting point, we can then use a template to generate test data, where the template specifies the transformations and test types to be applied. The template should explicitly state the hypothesis being used for a given test type. For example, the template might state the following hypothesis: 'applying rotation to a random sample of $p$\% of image patches results in images where $q$\% of the resulting images are allocated the same semantic label as the original image'. The user of the template would then specify $p$ and $q$ in order to generate the test images and check the hypothesis. There are many possibilities, for example, sweeping through the values of $p$ and $q$, or combining test types by applying rotation followed by occlusion. The Vision Checklist program can generate a large amount of rotated images based on the specified template. In figure~\ref{fig:tpro}, we demonstrate this process on a base image with different rotation rates. 
In Figure~\ref{fig:template}, we demonstrate a simplified user interface, which allows the user to specify the range of test types applied to the collection of test images.

\section{Visual Explanations}
\begin{figure}[t]
 \centering
   \subfloat[LIME and SHAP analysis on the adversarial example.\label{fig:shaplimeoriginal}]
   {\includegraphics[width=0.2\columnwidth]{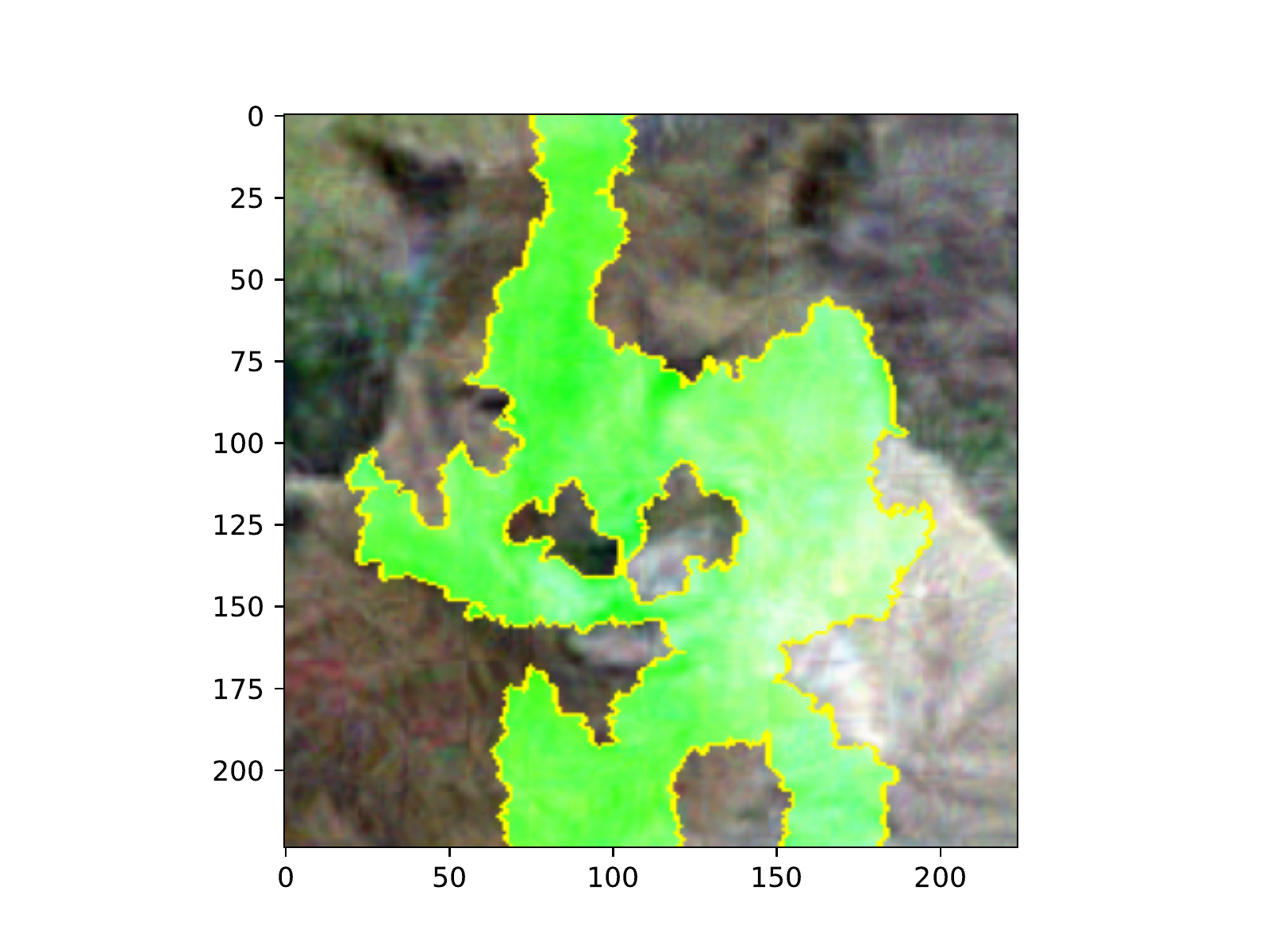}\includegraphics[width=0.25\columnwidth]{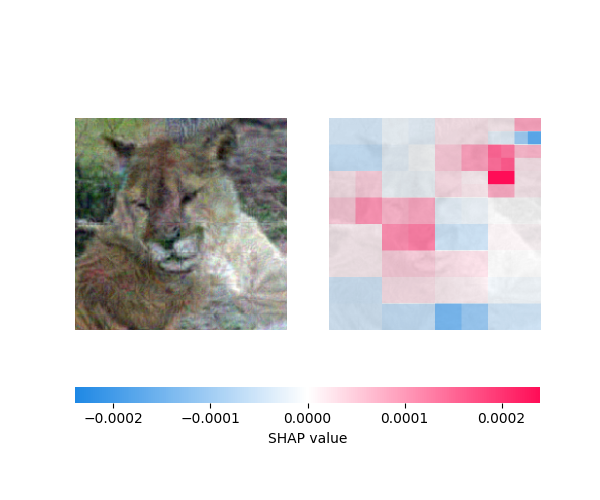}}
   \hspace{1mm}
   \subfloat[LIME and SHAP analysis on the rotated example.\label{fig:shaplimerotate}]
   {\includegraphics[width=0.2\columnwidth]{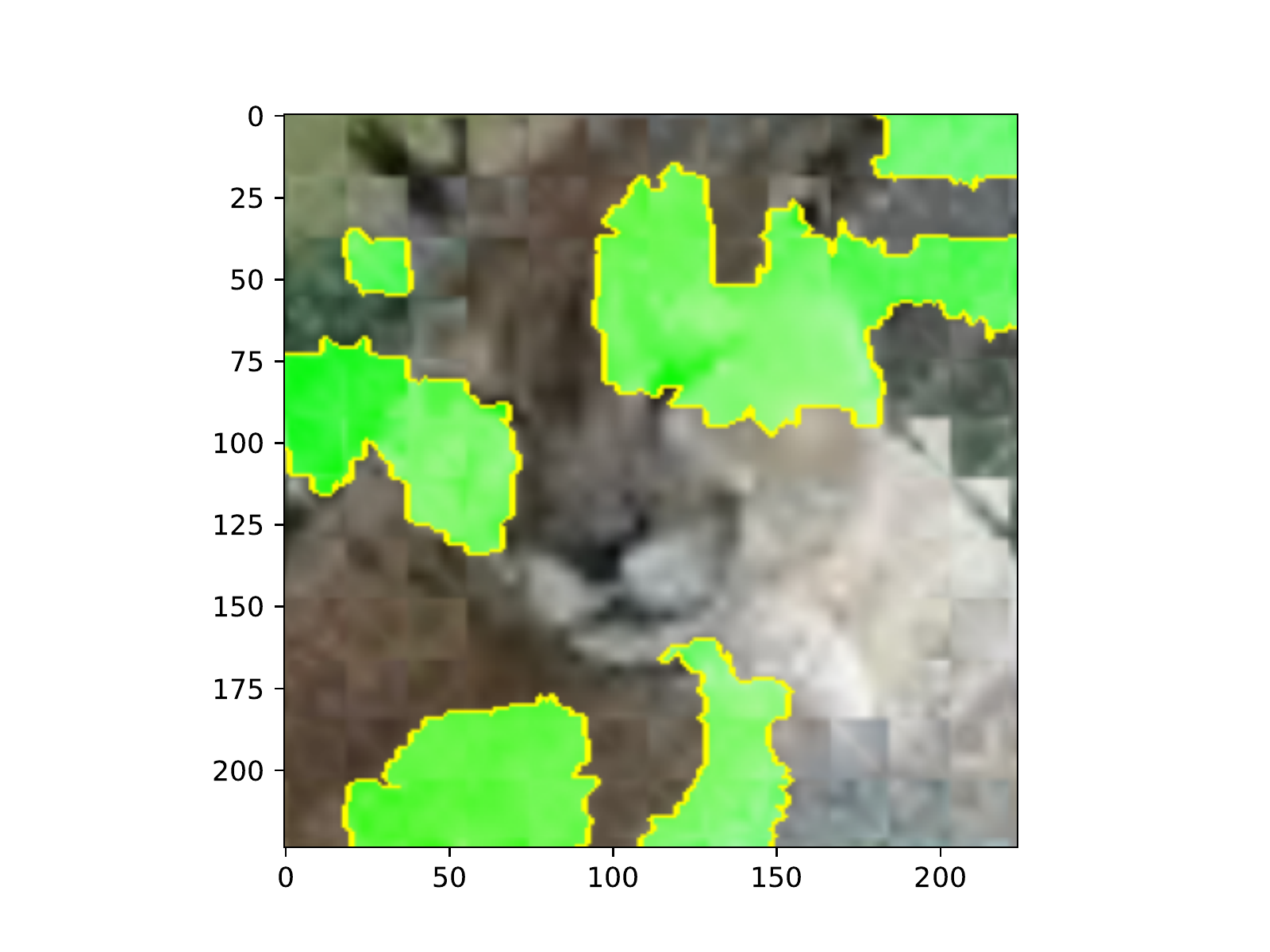}\includegraphics[width=0.25\columnwidth]{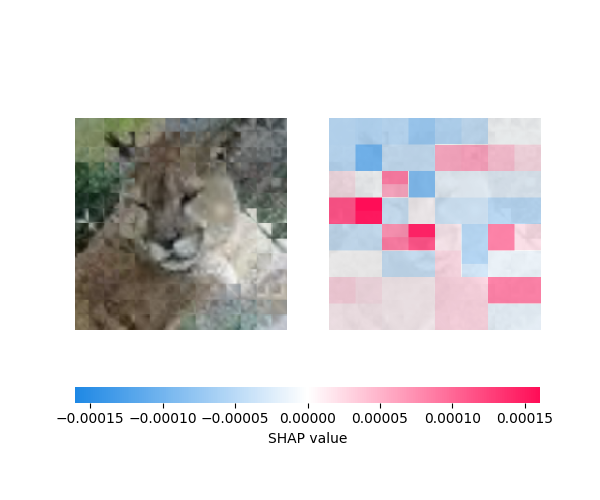}}
   \hspace{1mm}
   \subfloat[LIME and SHAP analysis on the occluded example.\label{fig:shaplimeocclude}]
   {\includegraphics[width=0.2\columnwidth]{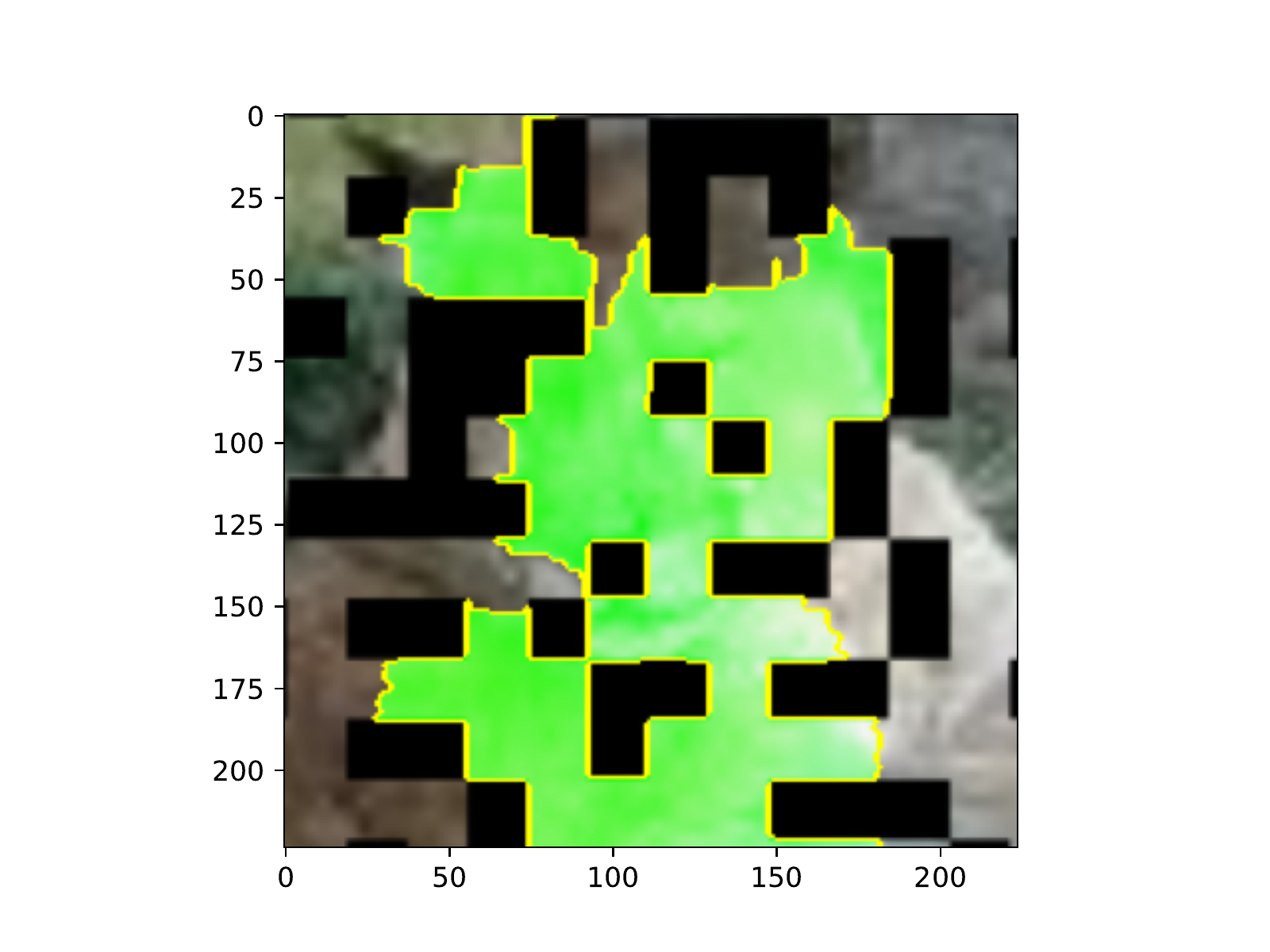}\includegraphics[width=0.25\columnwidth]{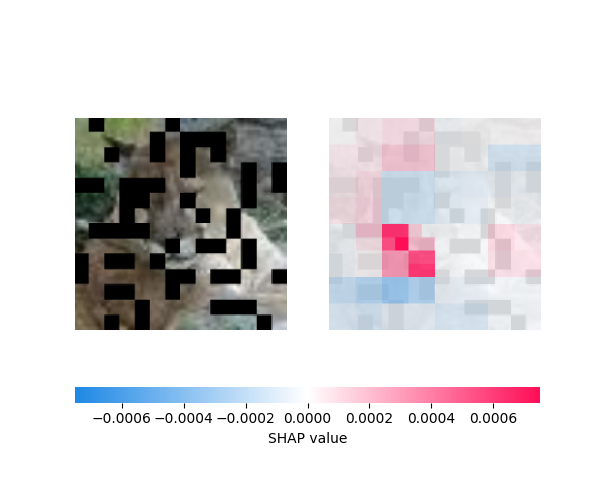}}
   \caption{Visual explanation test examples by LIME and SHAP.}
    \label{fig:shaplime}
\end{figure}

Our approach for visual explanations is best explained by an example.  Fig~\ref{fig:shaplimeoriginal} shows the original picture of a cougar in the centre. To its left, LIME~\cite{ribeiro2016should} highlights the interpretable components that contributed to the prediction, coloured in green. On the right, SHAP~\cite{NIPS2017_7062} computes the degree to which each feature contributed to the prediction, based on a game theoretic approach. The interpretable components, or features, of the image are displayed as coloured superpixels. Here, the more significant regions are shown in red, and less significant regions are shown in blue. The superpixels that SHAP and LIME highlight are associated with the shape of the cougar's muzzle and ears, as well as its coat. These features make intuitive sense; they are the features that a human would typically use to identify a cougar as well.

Fig~\ref{fig:shaplimerotate} shows the SHAP/LIME results on a test image generated by rotating outer patches, and fig~\ref{fig:shaplimeocclude} shows these results for a test image generated by occluding some patches. It is noticeable that the transformations have an impact on explanations; where, on the original image, the centre contributed to the prediction, all transformed image explanations have taken into account these transformations.

For the outer rotation example, LIME shows that the outer parts contribute more than the centre where the muzzle area is. SHAP, on the contrary, shows similar results compared to the original image, but attributes a higher contribution to the centre and middle left outer parts of the image, and disregards the shape of the ear on the upper right corner. For the random occlude transformation, the LIME results show that the black pixels are completely excluded from the explanation. SHAP seems to focus on the centre of the image again but, because of occlusion, the middle left part and the ear on the upper right corner are considered to have less of  a contribution to the prediction.

From these examples, we see that transformations have an impact on the results given by explainability tools like SHAP and LIME. The results show different superpixels contributing to the prediction depending on the transformation operator that has been applied. Overall, these explanations offer a rich tool to evaluate the model and help us understand how the transformations have impacted the model predictions.

\section{Experiments}
In this section, we present experiments using the Vision Checklist to evaluate a model in different scenarios. The main model we test is the Vision Transformer (ViT)~\cite{dosovitskiy2020image}. For comparison, we carry out the same experiments using Resnet50~\cite{he2016deep}. Specifically, we evaluate the models on a general image recognition task using the Tinyimagenet, CIFAR100 and CIFAR10 datasets. For the out-of-distribution robustness task, we employ the waterbirds dataset~\cite{koh2021wilds} for artificial confounding and the imagenet-A, imagenet-R and imagenet-C datasets for natural corruption~\cite{hendrycks2021nae,hendrycks2019robustness}. For the medical diagnosis task, we use the Camelyon17 challenge dataset and the Chexpert dataset~\cite{irvin2019chexpert}.

\subsection{Image recognition}
\begin{figure}[t]
 \centering
   \subfloat[Tinyimagenet \label{fig:vittinyimagenet}]
   {\includegraphics[width=0.3\columnwidth]{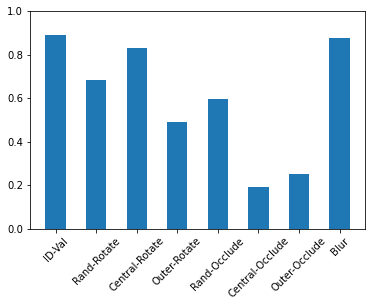}}
   \hspace{1mm}
   \subfloat[CIFAR100 \label{fig:vitcifar100}]
   {\includegraphics[width=0.3\columnwidth]{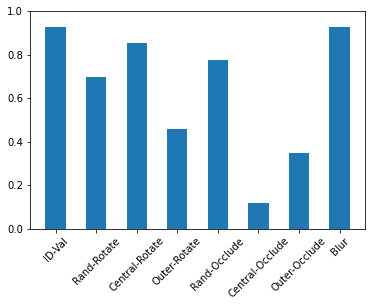}}
   \hspace{1mm}
   \subfloat[CIFAR10 \label{fig:vitcifar10}]
   {\includegraphics[width=0.3\columnwidth]{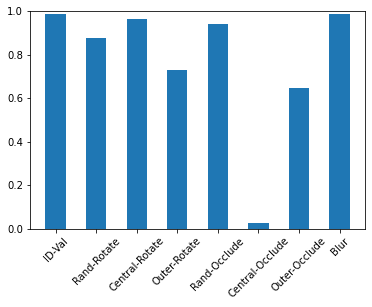}}
   \caption{Checklists for Vision Transformer. Y-axis represents the pass rate.}
    \label{fig:checkvit}
\end{figure}

\begin{figure}[]
 \centering
   \subfloat[Tinyimagenet \label{fig:resnettinyimagenet}]
   {\includegraphics[width=0.3\columnwidth]{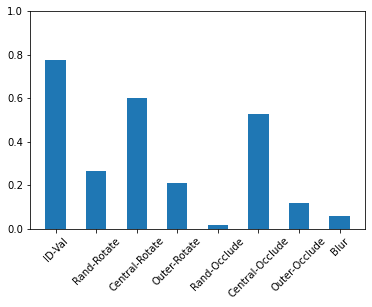}}
   \hspace{1mm}
   \subfloat[CIFAR100 \label{fig:resnetcifar100}]
   {\includegraphics[width=0.3\columnwidth]{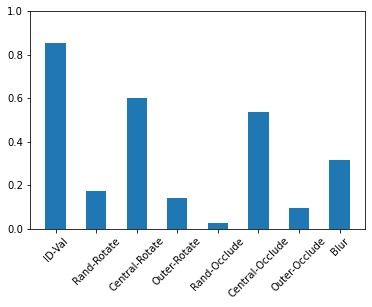}}
   \hspace{1mm}
   \subfloat[CIFAR10 \label{fig:resnetcifar10}]
   {\includegraphics[width=0.3\columnwidth]{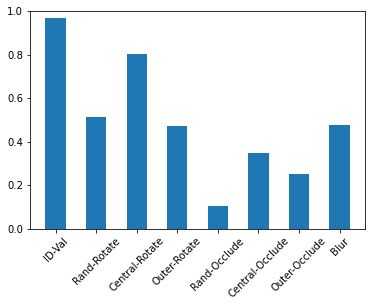}}
   \caption{Checklists for Resnet50. Y-axis represents the pass rate.}
    \label{fig:checkresnet}
\end{figure}

In this experiment, we use our Vision Checklist to evaluate ViT and Resnet50 on general image recognition tasks. The Vision Checklist generates test cases based on the Tinyimagenet, CIFAR100 and CIFAR10 datasets. The hypothesis we use for generating test samples can be described as follows: 'for a random patch rotate and occlude operation, the semantic meanings are kept from original ground truth images.' Then, the Vision Checklist tool generates a report with the prediction output of a given model and hypothesis used to generate test samples. In Figure~\ref{fig:checkvit} and \ref{fig:checkresnet}, we plot the test results for ViT and Resnet50 respectively on our chosen three datasets. The bar chart shows the proportion of images that are correctly labelled.  From the results, we can see that the patch based occlude operation has much more influence than the rotation and blur operations. The central patch based transformation has more of an effect on the model's performance, and this matches the hypothesis that 'the central region of an image indicates the semantic importance of the ground truth.' The results also show that due to the significant drop in performance from the central region perturbation, the model focused on the semantically important region, just as we hypothesized.

\begin{figure}[]
 \centering
   \subfloat[Vision Transformer \label{fig:vitro}]
   {\includegraphics[width=0.4\columnwidth]{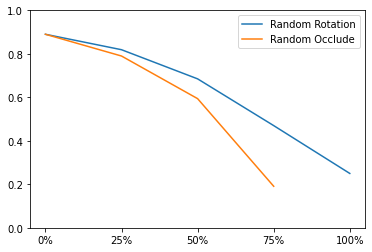}}
   \hspace{1mm}
   \subfloat[Resnet50 \label{fig:resnetro}]
   {\includegraphics[width=0.4\columnwidth]{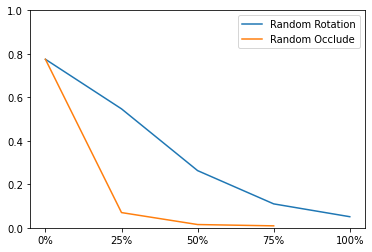}}
   \caption{Performance w.r.t percentage of perturbed patches on TinyImagenet dataset. Y-axis represents the pass rate.}
    \label{fig:pertpercent}
\end{figure}

Figure~\ref{fig:pertpercent}) demonstrates that Vision Transformer is more robust against patch-based transformations compared to Resnet50. This indicates that ViT learns features that can survive from those patch-based perturbations, while Resnet50 learns convolution features that are vulnerable under such operations. Similar results are also reported by~
\cite{qin2021understanding}. Understanding how features learned by models can survive under such patch-based perturbations is not trivial. This can help users design more robust models against distributional shifts and adversarial attacks.

In Figure~\ref{fig:f1g3}, we demonstrate cases of counterfactual analysis and adversarial analysis. In those cases, our Vision Checklist can generate counterfactual and adversarial examples by perturbing the images and keeping the semantic information.

\subsection{OOD robustness}
\begin{figure}[]
 \centering
   \subfloat[Waterbirds \label{fig:vitWaterbirds}]
   {\includegraphics[width=0.3\columnwidth]{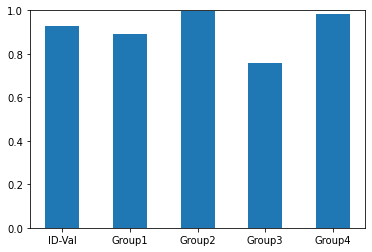}}
   \hspace{1mm}
   \subfloat[Natural Corruption \label{fig:vitnc}]
   {\includegraphics[width=0.3\columnwidth]{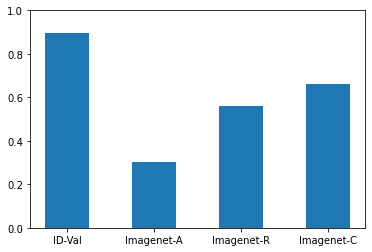}}
   \caption{OOD generalization for Vision Transformer. Y-axis represents the pass rate.}
    \label{fig:OODvit}
\end{figure}

\begin{figure}[]
 \centering
   \subfloat[Waterbirds \label{fig:resnetwaterbirds}]
   {\includegraphics[width=0.3\columnwidth]{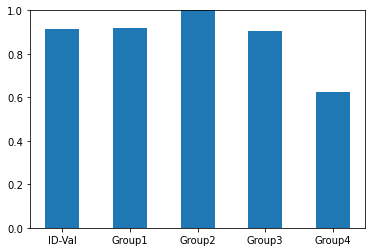}}
   \hspace{1mm}
   \subfloat[Natural Corruption \label{fig:resnetnc}]
   {\includegraphics[width=0.3\columnwidth]{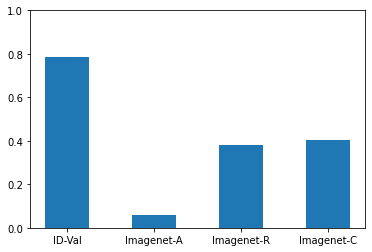}}
   \caption{OOD generalization for Resnet50. Y-axis represents the pass rate.}
    \label{fig:OODresnet}
\end{figure}

\begin{figure}[]
 \centering
   \subfloat[Common Yellow Throat original image \label{fig:cytori}]
   {\includegraphics[width=0.3\columnwidth]{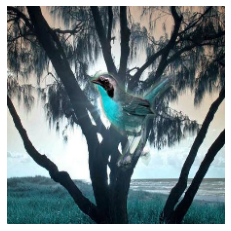}}
   \hspace{1mm}
   \subfloat[Common Yellow Throat attention region \label{fig:cytatt}]
   {\includegraphics[width=0.3\columnwidth]{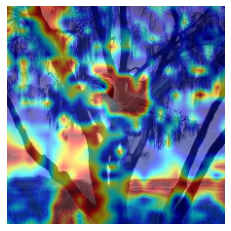}}
   \caption{Visual explanation for the attended region for the waterbirds dataset.}
    \label{fig:waterbirdatt}
\end{figure}

In this experiment, we investigate the test types that can demonstrate a model's out-of-distribution robustness. By out-of-distribution, we mean that the test data are generated from a distribution which is different from the training data. In some cases we may not know the distribution of the training data but we can consider transformations that result in a distribution of test data that is quite different from the training data.
There are a lot of definitions of out-of-distribution. In this paper, the OOD data that we consider are two-fold: on the one hand, we consider artificially generated OOD data based on an assumed generative model; on the other hand, we consider corruptions and perturbations that would not affect the semantic meaning of the images. 
For artifact OOD, we assume that the input images consist of core features combined with spurious features. Core features are groups of pixels that are associated with the semantic meaning of the ground truth, e.g. the shape of the object in the image. Spurious features are groups of pixels that are correlated with the semantic label but will change when the environment changes, e.g. the brightness or quality of the image, hair color or posture of the person in the image. 
We assume a model for the data generating process. In this model, we assume that the spurious features are affected both by their semantic label and environment label. When the environment changes, spurious features will change. Hence, the aim of this test is to check whether the model can learn core features that will survive environment shifts. The test cases used in this experiment, using the waterbird datasets~\cite{koh2021wilds}, are generated by artificially replacing the backgrounds of images to create spurious correlations between the background and the semantic label. In Figure~\ref{fig:vitWaterbirds} and \ref{fig:resnetwaterbirds}, we plot the performance of the model by grouping the test data based on label and background types. As we can see, even though the average performance is high, the worst-case group performance is relatively low due to the OOD generalization problem. In Figure~\ref{fig:waterbirdatt}, we show visual explanation results for the waterbirds datasets using the ViT model. As we can see from the figure, the model relies heavily on background information to make predictions. This could lead to spurious correlation with the environment, indicating vulnerabilities to environmental shifts in the test cases.

To validate the robustness of the model against natural corruption and perturbations, we employ the imagenet-A,R,C datasets~\cite{hendrycks2021nae,hendrycks2019robustness} for extended test cases. In Figure~\ref{fig:vitnc} and \ref{fig:resnetnc}, we demonstrate the test results. As we can see, Vision Transformer is more robust than Resnet50 against distribution shifts and corruptions.

\subsection{Medical diagnosis}
One goal of our Vision Checklist is to better understand how the performance of the model changes in response to changing environments. This is especially important for high risk tasks such as medical diagnosis. In this experiment, we used the Vision Checklist to build test cases for medical datasets.

\begin{figure}[t]
 \centering
   \subfloat[Chexpert \label{fig:vitchexpert}]
   {\includegraphics[width=0.32\columnwidth]{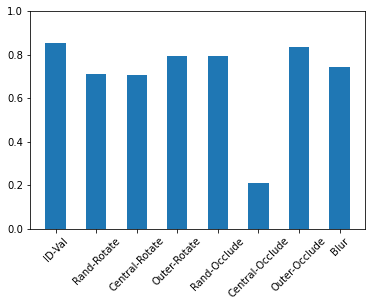}}
   \hspace{1mm}
   \subfloat[Camelyon17 \label{fig:vitcamelyon17}]
   {\includegraphics[width=0.32\columnwidth]{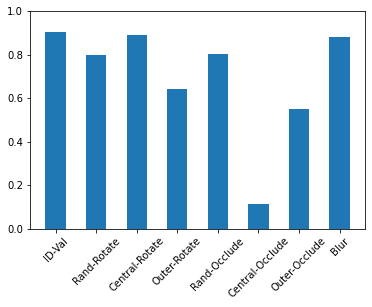}}
   \caption{Medical diagnosis for Vision Transformer. Y-axis represents the pass rate.}
    \label{fig:mdvit}
\end{figure}

\begin{figure}[t]
 \centering
   \subfloat[Chexpert \label{fig:resnetchexpert}]
   {\includegraphics[width=0.32\columnwidth]{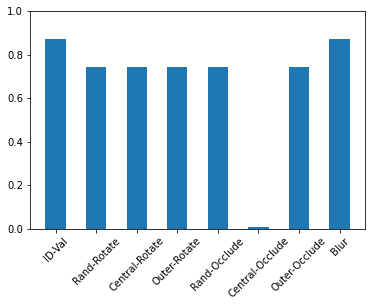}}
   \hspace{1mm}
   \subfloat[Camelyon17 \label{fig:resnetcamelyon17}]
   {\includegraphics[width=0.32\columnwidth]{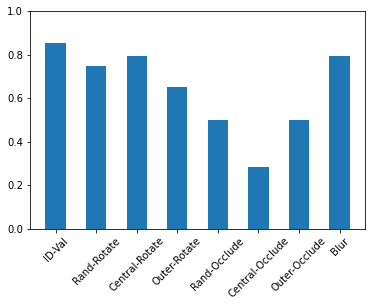}}
   \caption{Medical diagnosis for Resnet50. Y-axis represents the pass rate.}
    \label{fig:mdresnet}
\end{figure}

\begin{figure}[t]
 \centering
   \subfloat[Original Image \label{fig:oricamelyon17}]
   {\includegraphics[width=0.2\columnwidth]{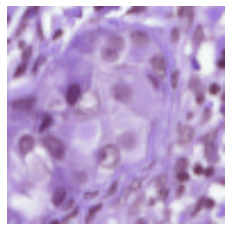}}
   \hspace{1mm}
   \subfloat[Counterfactual Generation \label{fig:cfcamelyon17}]
   {\includegraphics[width=0.2\columnwidth]{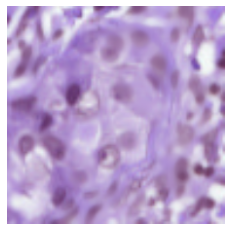}}
   \subfloat[Attention Region \label{fig:attregion}]
   {\includegraphics[width=0.2\columnwidth]{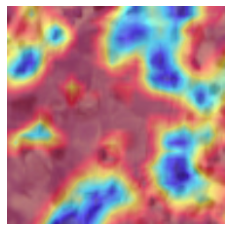}}
   \caption{Counterfactual analysis for medical diagnosis.}
    \label{fig:cfana}
\end{figure}

The datasets we use to generate test cases are from the Chexpert paper~\cite{irvin2019chexpert} and the Camelyon17 challenge~\cite{litjens20181399}. As shown in Figure~\ref{fig:mdvit} and \ref{fig:mdresnet}, the models are very sensitive to test cases generated by the central occlude operation. This supports the hypothesis that central regions are important regions for the semantic label, especially for some medical imaging datasets. In Figure~\ref{fig:cfana}, we conduct visual explanation on the counterfactual analysis. In the counterfactual examples, generated following~\cite{chang2018explaining}, the regions that contribute most to the model's prediction are blurred with infilling algorithms automatically. 

\section{Building A Model Card Using Vision Checklist}
Our Vision Checklist provides a useful set of evaluation criteria for vision models, which would fit nicely under the “evaluation” section of the model card proposed by Mitchell, et al~\cite{mitchell2019model}. The checklist enhances the transparency provided by a model card by providing a specific set of robustness checks, and explicitly stating the hypotheses and test types used in the evaluation process. This increased transparency allows users of the model to make better informed decisions regarding when the model can be appropriately used. With the help of domain experts, the Vision Checklist could also be expanded to integrate domain expertise and include domain-specific checks. And while the Vision Checklist is only for vision models, it does demonstrates the usefulness of coming up with similar templates for other model types, such as language models or videos.

\subsection{Implications for Policy and Governance}
From a policy and regulatory point of view, a model card using the Vision Checklist could provide an opportunity to communicate the error rates of vision models to different audiences. In a field like computer vision, which is typically only accessible to a specific technical audience, a model card could be used as a “boundary object”~\cite{star_institutional_1989}. In Science and Technology Studies literature, Star and Griesemer (1989) constructed the idea of a “boundary object”, to help translate technical concepts and find a common language across different disciplines~\cite{star_institutional_1989}. The model card could be seen as a type of boundary object,  and it can be used to establish commitments across stakeholders including (but not limited to) designers, deployers, and regulators. 

The Vision Checklist could also help regulators understand error profiles and set acceptable standards and criteria for evaluating vision models. Having a regularly updated model card for vision models could also be useful as an ongoing tool for governance of systems utilizing machine learning models. As the MHRA~\cite{noauthor_software_nodate} and FDA ~\cite{health_artificial_2021} guidance shows, policymakers and regulators have an interest in both "pre-market" and "post-market" continuous evaluation of systems. For those involved in safety assurance~\cite{mcdermid2019towards}, model cards with the Vision Checklist would provide particularly valuable evidence. 

While tools for increasing transparency and translation across domains would aid in the governance of systems utilizing machine learning models, they are only part of the picture. Model cards and evaluation tools like Vision Checklist help with notions of forward-looking governance and the setting of standards for performance and uncertainty quantification. They also may prove useful for regulators in the process of approving a given system for widespread adoption~\cite{health_artificial_2021}. However, they do not necessarily provide guardrails in the case where an approved system harms an individual or community. Governance of this kind on an ongoing basis requires that responsibility is taken for the outcomes of algorithmic and autonomous systems. While domains like medicine have defined agencies like the MHRA in the UK and the FDA in the US to protect patient interests, there are high risk applications of machine learning models in domains that are largely unregulated, such as automated hiring~\cite{ajunwa2021auditing}. In these cases, there would still be unanswered questions regarding who the ultimate arbiter would be of whether a particular use of a model was appropriate or whether the error modes could be considered acceptable or not.

Even if there is a central regulatory agency overseeing the evaluation of machine learning models in a given domain area, the "many hands" problem, or complexity of assigning responsibility to any one individual when there are so many different engineers, designers, data scientists involved,  still remains unresolved~\cite{davis_aint_2012}. And while the the existence of a boundary object like a model card would create information parity across stakeholders, it is still unclear how responsibility for the outcome of the model would ultimately be assigned or taken, given the number of involved parties and their differing priorities. This "responsibility gap"~\cite{santoni_de_sio_four_2021}, or lack of moral or legal accountability for harms, is exacerbated by organisations increasingly using technology that is not even created or tested in house. There are also added complexities around the allocation of responsibility to individuals versus organisations, and what that entails from a governance standpoint. And finally, even if the matter of who should be held responsible is settled, it is still unclear who exactly the responsible parties are answerable to, and what those who are impacted by the outcomes of a model are ultimately owed from a moral and legal standpoint. These questions remain largely unanswered in the literature around governance of trustworthy autonomous systems. Our future work will look to address these questions.


\section{Conclusion}
Model evaluation is becoming a more important problem for the current machine learning community, as stakeholders become increasingly concerned about model performance in unknown deployment environments. Due to the high uncertainty in deployment environments, measures based on a small set of hold-out data are not enough for model evaluation. In this paper, we provide guidance on the evaluation of image recognition models, applied to domains like autonomous driving and medical diagnosis. Firstly, we define several properties of good model evaluation systems. We suggest that hypothesis definition plays an important role in the evaluation process, especially for the reproducibility of test results. Then we propose that in order to test the model's capabilities from multiple aspects, it is necessary to use the entire dataset as potential candidates for generating test cases. Finally, we suggest that a good model evaluation system should ensure identifiability, which is essential for the replicability and interpretability of a model. Guided by these properties we propose a framework called the Vision Checklist. The Vision Checklist focuses on building transformations based on image patches. By defining several types of patch transformations and test types with different hypotheses, the Vision Checklist provides templates for generating more comprehensive test cases automatically. The checklist tool provides quantitative analysis summarizing the error modes of the model's performance on those generated test cases. The checklist can also provide visual explanations with adversarial analysis and counterfactual analysis to help users identify problems in the decision making process and generate better decision outcomes.

\section{Acknowledgement}
This work was supported by a grant from the UKRI Strategic Priorities Fund to the UKRI Research Node on Trustworthy Autonomous Systems Governance and Regulation (EP/V026607/1, 2020-2024).
\bibliographystyle{ACM-Reference-Format}
\bibliography{ViTChecklist.bib}










\end{document}